\documentclass[11pt]{article}

\usepackage[final]{acl}

\usepackage{times}
\usepackage{latexsym}

\usepackage[T1]{fontenc}

\usepackage[utf8]{inputenc}

\usepackage{microtype}

\usepackage{inconsolata}

\usepackage{graphicx}

\usepackage{algorithm}
\usepackage{algorithmic}
\usepackage{booktabs}
\usepackage{multirow}
\usepackage{amsmath}
\usepackage{xcolor}
\definecolor{deepgreen}{RGB}{0,140,0}   
\definecolor{deepred}{RGB}{220,0,0}     
\usepackage{tabularx}
\usepackage{bm}   

\title{The Bidirectional Process Reward Model}

\author{
 \textbf{Lingyin Zhang\textsuperscript{1,2}},
 \textbf{Jun Gao\textsuperscript{3}},
 \textbf{Xiaoxue Ren\textsuperscript{3,4}},
 \textbf{Ziqiang Cao\textsuperscript{1,2}\thanks{~Corresponding author.}}
\\
 \textsuperscript{1}School of Computer Science and Technology, Soochow University\\
 \textsuperscript{2}Biomedical Basic Research Center of Jiangsu, Soochow University, Suzhou, Jiangsu 215123, China\\
 \textsuperscript{3}School of Software Technology, Zhejiang University\\
 \textsuperscript{4}Hangzhou High-Tech Zone (Binjiang) Institute of Blockchain and Data Security
\\
 \small{
   \href{mailto:lyzhanglyzhang@stu.suda.edu.cn}{lyzhanglyzhang@stu.suda.edu.cn, }
   \href{mailto:jgao1106@zju.edu.cn}{jgao1106@zju.edu.cn, }
   \href{mailto:xxren@zju.edu.cn}{xxren@zju.edu.cn, }
   \href{mailto:zqcao@suda.edu.cn}{zqcao@suda.edu.cn}
 }
}

\begin{document}
\maketitle
\begin{abstract}
Process Reward Models (PRMs), which assign fine-grained scores to intermediate reasoning steps within a solution trajectory, have emerged as a promising approach to enhance the reasoning quality of Large Language Models (LLMs).
However, most existing PRMs rely on a unidirectional left-to-right (L2R) evaluation scheme, which restricts their utilization of global context.
In light of this challenge, we propose a novel bidirectional evaluation paradigm, named \textbf{Bi}directional \textbf{P}rocess \textbf{R}eward \textbf{M}odel (\textbf{BiPRM}).
BiPRM incorporates a parallel right-to-left (R2L) evaluation stream, implemented via prompt reversal, alongside the conventional L2R flow.
Then a gating mechanism is introduced to adaptively fuse the reward scores from both streams to yield a holistic quality assessment.
Remarkably, compared to the original PRM, BiPRM introduces only a 0.3\% parameter increase for the gating module, and the parallel execution of two streams incurs merely 5\% inference time latency.  
Our extensive empirical evaluations spanning diverse benchmarks, LLM backbones, PRM objectives and sampling policies demonstrate that BiPRM consistently surpasses unidirectional baselines, achieving an average relative gain of 10.6\% over 54 solution-level configurations and 37.7\% in 12 step-level error detection scenarios.
Generally, our results highlight the effectiveness, robustness and general applicability of BiPRM, offering a promising new direction for process-based reward modeling.\footnote{~\url{https://github.com/LingyinZhang/BiPRM}}
\end{abstract}

\begin{figure*}[!t]
\centering
\includegraphics[width=0.99\linewidth]{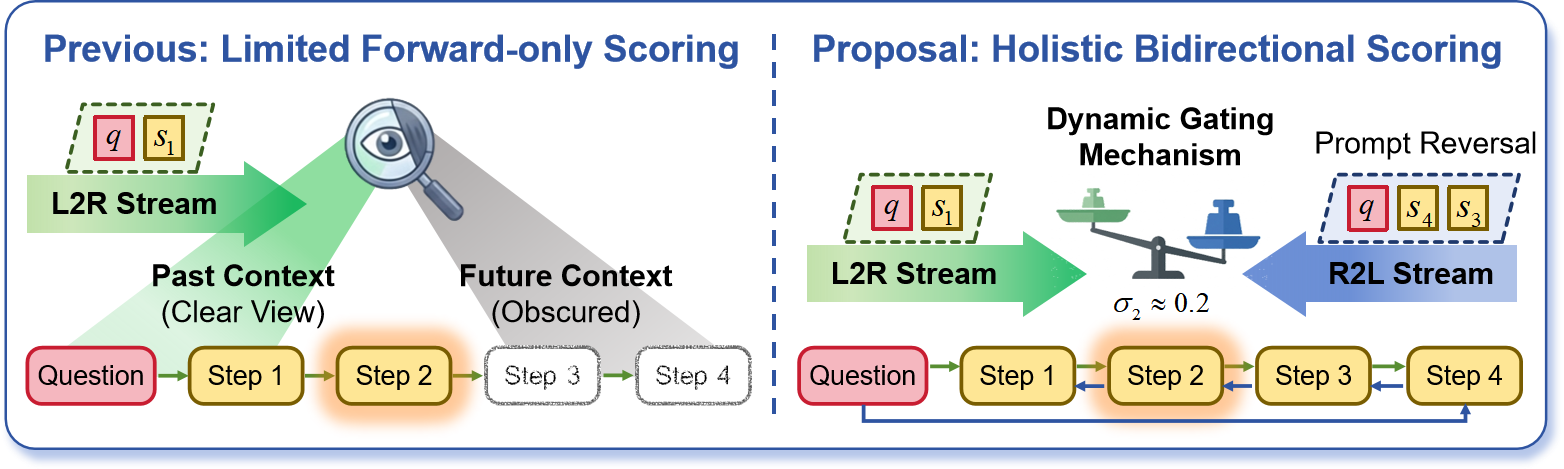}
\caption{Comparison of evaluation paradigms when scoring \textbf{Step 2}. (Left) Conventional unidirectional PRMs are restricted to past context, lacking access to subsequent steps required for verification. (Right) BiPRM integrates a parallel R2L stream to incorporate future context, enabling effective retrospective verification of the current step.}
\label{fig:Architecture comparison}
\end{figure*}

\section{Introduction}
Process reward models (PRMs) aim to augment the response quality of policy models by granularly scoring a sequence of intermediate steps~\cite{lightman2023let}.
Currently, PRMs have been widely applied to align reinforcement learning (RL) models with human preferences~\cite{lightman2023let}, as well as to support Test-Time Scaling (TTS)~\cite{jaech2024openai,guo2025deepseek} strategies in a spectrum of complex cognitive tasks such as mathematical reasoning~\cite{lightman2023let,li2024process,zhang2024rest,xia2025evaluating,ma2025step}, code generation~\cite{le2022coderl,li2024ircoco} and question answering~\cite{carta2022eager,bi2023boosting}.

Existing studies typically formulate PRM training via various objectives, ranging from classification tasks using binary cross-entropy (BCE)~\cite{wang2024math,shao2024deepseekmath,luo2024improve,lightman2023let} to regression tasks employing mean squared error (MSE)~\cite{zhang2024rest,wang2024q} or Q-value rankings loss~\cite{li2024process} for more fine-grained supervision. 
Since PRMs are typically built upon generative policy models and share the same parameter initialization, existing works predominantly adopt a unidirectional left-to-right (L2R) evaluation paradigm, where step plausibility is assessed sequentially based on preceding context.
However, this paradigm inherently restricts access to global context (Figure~\ref{fig:Architecture comparison}, Left), which is critical when an early step's validity hinges on downstream consequences. For instance, in the mathematical induction case shown in Table~\ref{tab:case study}, determining the correctness of the hypothesis in Step 2 necessitates verifying the derivation logic in Steps 3 and 4. 
By relying exclusively on past context, standard PRMs miss these essential backward signals, thereby constraining their global optimization capability~\cite{liu2024bi}.
Although recent studies such as BiRM~\cite{chen2025better} attempt to incorporate future guidance via an auxiliary value head, they remain fundamentally forward-looking predictors, without performing direct retrospective verification through structural context reversal.

To address this limitation, we propose the Bidirectional Process Reward Model (BiPRM), a novel evaluation paradigm for PRMs that draws inspiration from the architecture of Bidirectional Long Short-Term Memory (BiLSTM) networks~\cite{graves2012long}. 
As shown in Figure~\ref{fig:Architecture comparison} (Right), BiPRM synergizes a parallel R2L evaluation stream with the conventional L2R flow, allowing subsequent steps to provide retrospective evidence for earlier ones. 
Noting that increasing context length is generally associated with improved reliability of reasoning (Figure~\ref{fig:mae_sigma_step}a), we adopt a dynamic gating mechanism, rather than static averaging, to integrate scores from the two streams.
Since the R2L stream is efficiently realized by simply reversing the reasoning trajectory via prompt modifications, BiPRM requires only a 0.3\% parameter increase for the gating module. 
Through parallel execution, BiPRM effectively mitigates the computational overhead of dual-stream evaluation, incurring only an approximate 5\% increase in wall-clock latency while significantly enhancing verification accuracy.

We comprehensively conduct experiments on two solution-level benchmarks (GSM-Plus~\cite{li2024gsm} and MATH500~\cite{hendrycks2021measuring}) and one diagnostic step-level benchmark (ProcessBench~\cite{zheng2025processbench}).
Evaluating our method across three backbones of varying scales and three PRM objectives, experimental results demonstrate that BiPRM consistently outperforms unidirectional baselines.
Specifically, BiPRM achieves an average relative gain of 10.6\% across 54 solution-level configurations and 37.7\% across 12 step-level error detection scenarios.
These consistent improvements confirm that the bidirectional evaluation paradigm provides a robust modeling advantage regardless of model capacity or training objective.

In summary, our contributions are as follows:
\begin{itemize}
    \item  As far as we know, we are the first to propose the concept of the bidirectional evaluation paradigm for PRMs, addressing unidirectional models' limitations in local perspective through dual-stream fusion.
    \item BiPRM achieves an excellent performance-efficiency trade-off, adding only about 5\% inference latency by running the R2L stream in parallel.
    \item As a general framework, BiPRM can adapt to diverse PRM architectures, offering new directions for future process reward modeling.
\end{itemize}

\begin{figure*}[!ht]
\vspace{0.5em}
    \centering    
    \includegraphics[width=0.9\linewidth]{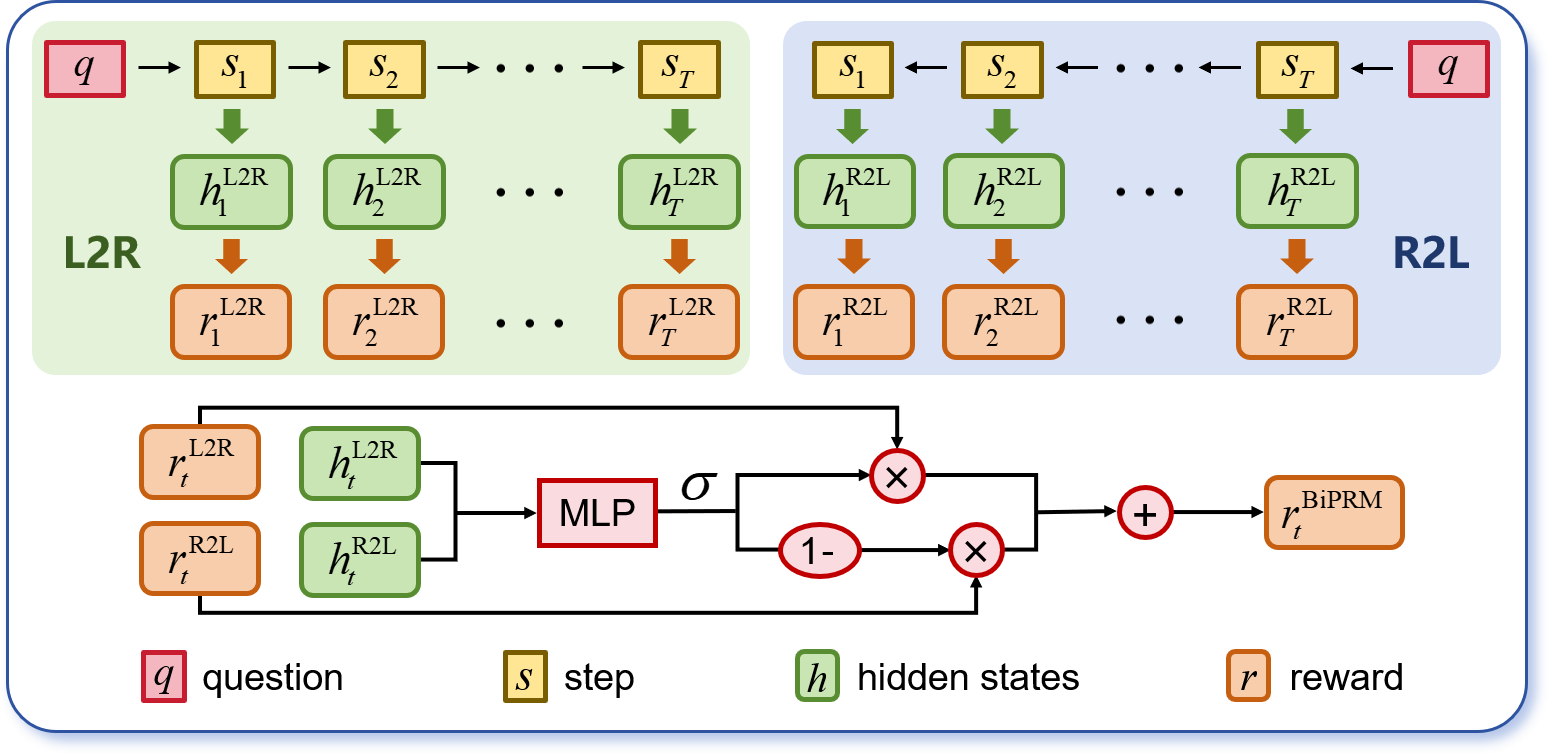}
    \caption{Overview of the BiPRM Architecture. The model synthesizes parallel L2R and R2L evaluation flows via a dynamic gating mechanism for step-wise reward modeling.}
    \label{fig:Comparison chart}
\vspace{0.2em}
\end{figure*}

\section{Related work}
\subsection{Process Reward Models}
Process Reward Models (PRMs) improve the alignment of language models with human reasoning preferences by providing fine-grained feedback at intermediate reasoning steps rather than evaluating only final answers~\cite{lightman2023let}.
Early works such as PRM800K~\cite{lightman2023let} utilized large-scale human annotations to train token-level step classifiers.
Subsequent approaches have explored automated supervision to reduce annotation costs, estimating step quality via Monte Carlo rollouts~\cite{wang2024math,wang2024multi} or modeling structured trajectories using Monte Carlo Tree Search~\cite{luo2024improve,zhang2024rest}.
Recent advancements include ranking-based objectives~\cite{li2024process} and advantage-based reward definitions~\cite{lu2024autopsv,setlur2024rewarding}.
Despite these innovations, most existing PRMs adhere to a strict L2R evaluation paradigm.
This unidirectional constraint inherently limits the model's ability to leverage global context for verifying the consistency of the entire trajectory.
Moreover, the increasing trajectory complexity introduced by emerging multimodal contexts~\cite{gao2025interleaved} and efficient in-context learning~\cite{gao2025uniicl, gao2025aim} exacerbates the blind spots of unidirectional PRMs, further underscoring the necessity of our bidirectional framework.

\subsection{Order of Reasoning}
The order of reasoning has been shown to significantly influence model performance.
While early studies suggested a symmetry between forward and backward prediction~\cite{shannon1951prediction}, recent work indicates that forward generation typically yields lower perplexity~\cite{papadopoulos2024arrows}.
However, non-monotonic and bidirectional approaches have shown promise in specific domains.
For instance, reversing input sequences has proven effective in machine translation~\cite{sutskever2014sequence}, and bidirectional encoders like BERT~\cite{devlin2019bert} excel at capturing contextual dependencies.
In reasoning tasks, backward chaining can sometimes outperform forward deduction~\cite{vinyals2015order, pfau2023eliciting}.
Furthermore, diffusion-based models~\cite{zhang2023planner,li2022diffusion,gong2024scaling} demonstrate the efficacy of planning in both directions.
Building on these insights, we propose integrating a backward verification stream 
into reward modeling to leverage the complementary strengths of bidirectional reasoning.

\section{Methodology}
\subsection{Preliminary}
Given a mathematical problem $q$, a policy language model generates a reasoning trajectory to solve the task.
This trajectory, often referred to as a Chain-of-Thought~\cite{wei2022chain}, is formally represented as $\tau=(q,\{s_1,...,s_T\})$, where $s_t$ denotes the $t$-th reasoning step and $s_T$ contains the final answer.
Standard PRMs typically adopt a unidirectional L2R evaluation paradigm, where each step $s_t$ is scored sequentially based on the question and prior context. 
The L2R reward function for step $s_t$ is expressed as
\begin{equation}
r_t^{\text{L2R}} = f_\theta(s_t \mid q, s_{<t}),
\end{equation}
where $f_\theta$ denotes the reward function parameterized by $\theta$.
Under this formulation, the reward $r_t^{\text{L2R}}$ is strictly independent of future steps $s_{>t}$, leading to the following gradient property:
\begin{equation}
\frac{\partial r_t^{\text{L2R}}}{\partial s_{t+k}} = 0 \quad \forall k \in [1, T - t].
\end{equation}
This zero gradient prevents future signals from propagating backward to facilitate retrospective verification, thereby restricting standard PRMs to a local perspective that lacks global consistency.

\subsection{Implementation of BiPRM}
To address the aforementioned limitation, we draw inspiration from BiLSTM and develop BiPRM, a bidirectional process reward model that incorporates a backward evaluation signal in parallel with the conventional L2R scoring path.
As illustrated in Figure~\ref{fig:Comparison chart}, BiPRM constructs a backward trajectory by logically inverting the reasoning sequence to $(q, \{s_T, s_{T-1}, \dots, s_1\})$.
This step-level reversal mimics the human cognitive process of retrospective verification, checking derivations from conclusions back to premises.
Consequently, the R2L reward function for a specific step $s_t$ is defined as
\begin{equation}
    r_t^{\text{R2L}} = f_\theta(s_t \mid q, s_{>t}).
\end{equation}
Notably, this R2L stream is implemented entirely via prompt reversal with no architectural modifications to the underlying LLM, requiring only a negligible 0.3\% parameter increase for the gating module.

To synthesize the bidirectional signals, BiPRM employs a step-wise dynamic gating mechanism.
Our empirical analysis (Section~\ref{sec:analysis}) reveals that L2R and R2L streams possess complementary error profiles, where L2R excels in later stages while R2L is more effective early on.
To exploit this position-dependent synergy, we compute a dynamic weight $\sigma_t$ for each step $t$ using a lightweight MLP:
\begin{equation}
    \sigma_t = \text{Sigmoid}\left(\text{MLP}\left([{h}_t^{\text{L2R}}; {h}_t^{\text{R2L}}]\right)\right),
\end{equation}
where $h_t^{\text{L2R}}$ and $h_t^{\text{R2L}}$ are the hidden states from the respective streams.
The final bidirectional reward for each step $t$ is computed as:
\begin{equation}
    r_t^{\text{BiPRM}} = \sigma_t \cdot r_t^{\text{L2R}} + (1 - \sigma_t) \cdot r_t^{\text{R2L}}.
\end{equation}
This bidirectional formulation ensures that the evaluation of each step is informed by the global context, allowing future information to influence current scoring:
\begin{equation}
    \frac{\partial r_t^{\text{BiPRM}}}{\partial s_{t+k}} = \sigma_t \frac{\partial r_t^{\text{L2R}}}{\partial s_{t+k}} + (1-\sigma_t) \frac{\partial r_t^{\text{R2L}}}{\partial s_{t+k}} \neq 0.
\end{equation}

Finally, to obtain the trajectory-level reward, we apply a reduction operation over the sequence of stepwise scores using an aggregation operator $\bigoplus \in \{\prod, \min, \max, \text{mean}\}$.
While a comprehensive sensitivity analysis of these different operators is provided in Appendix~\ref{sec:aggregation_analysis}, we adopt the minimum operator following previous works~\cite{li2024process, wang2024math}. 
This strategy is grounded in the logical "weakest link" principle, which encourages the model to filter out trajectories containing even a single fatal error. 
The overall trajectory reward is then given by
\begin{equation}
    R_{\text{BiPRM}}(\tau) = \min_{t=1}^T r_t^{\text{BiPRM}}.
\end{equation}
Due to the parallel execution of the two evaluation streams, BiPRM enhances verification effectiveness while incurring only an approximate 5\% increase in inference time latency.

\begin{table*}[!ht]
\centering
\small
    \begin{tabular}{ccccc}
    \toprule
    \multicolumn{1}{c@{}}{\textbf{Datasets}} & \multicolumn{2}{c}{\textbf{Source}} & \textbf{Quantity} & \multicolumn{1}{c@{}}{\textbf{Avg. Steps}} \\
    \midrule
    Training & \multicolumn{2}{c}{\multirow{2}{*}{\shortstack{Math-Shepherd~\cite{wang2024math}}}} & 397,927 & \multirow{2}{*}{6.3} \\
    Validation & & & 20,944 & \\
    \midrule
    \multirow{7}{*}{Evaluation} & \multirow{3}{*}{\shortstack{GSM-Plus\\~\cite{li2024gsm}}} & \shortstack{MetaMath-Mistral-7B~\cite{yu2024metamath}} & \multirow{3}{*}{\shortstack{2400 questions $\times$\\128 trajectories}} & 3.3 \\
    & & \shortstack{MuggleMath-13B~\cite{li2024query}} & & 3.2 \\
    & & \shortstack{Llama-3-70B-Instruct~\cite{grattafiori2024llama}} & & 2.9 \\
    \cmidrule(lr){2-5}
    & \multirow{3}{*}{\shortstack{MATH500\\~\cite{hendrycks2021measuring}}} & \shortstack{MetaMath-Mistral-7B~\cite{yu2024metamath}} & \multirow{3}{*}{\shortstack{500 questions $\times$\\128 trajectories}} & 3.9 \\
    & & \shortstack{MuggleMath-13B~\cite{li2024query}} & & 2.6 \\
    & & \shortstack{Llama-3-70B-Instruct~\cite{grattafiori2024llama}} & & 2.9 \\
    \cmidrule(lr){2-5}
    & \multicolumn{2}{c}{ProcessBench~\cite{zheng2025processbench}}  & 3400 & 7.2 \\
    \toprule
    \end{tabular}
    \caption{Statistical information of the training, validation and evaluation datasets.}
    \label{tab:datasets}
\end{table*}

\subsection{Training Objectives}
\label{sec:objectives}
Since the bidirectional evaluation framework we proposed does not change the underlying architecture of PRMs, this paper tests the following three existing objectives.
The first is the \textbf{Binary Cross-Entropy (BCE)} loss, which treats reward prediction as a classification problem.
It is defined as follows:
\begin{equation}
\mathcal{L}_\text{BCE}(\tau)={-\displaystyle \frac{1}{T}} \sum\limits_{t=1}^T \bigl( r_t \ln(\hat{r_t})+(1-r_t) \ln(1-\hat{r_t}) \bigr),
\end{equation}
where $r_t$ is the gold classification label of $t$-th step and  $\hat{r_t}$ is the predicted reward.
This loss is widely used due to its simplicity and alignment with classification-based annotations.

The second objective is the \textbf{Mean Squared Error (MSE)} loss, formulated as
\begin{equation}
\mathcal{L}_\text{MSE}(\tau)={-\displaystyle \frac{1}{T}} \sum\limits_{t=1}^T \bigl( \hat{r_t}-r_t \bigr)^2,
\end{equation}
which penalizes large deviations between the predicted and target reward values.
While MSE allows for finer-grained supervision, it is more sensitive to outliers in label distributions.

The third objective follows the \textbf{Q-value rankings loss} proposed in recent work~\cite{li2024process}, formulated as
\begin{align}
\Sigma_t
&= \sum_{q=0}^{t}\exp(\hat r_{cq}) + \sum_{w\in W}\exp(\hat r_w+\zeta), \notag
\\
\mathcal{L}_{\mathrm{Q}}(\tau)
&= -\frac{1}{|C|}\sum_{t=1}^{|C|}
   \log\frac{\exp(\hat r_{ct})}{\Sigma_t},
\end{align}
where $C$ and $W$ respectively denote the index lists of correct and incorrect steps in this trajectory, $|\cdot|$ denotes the length of this list, and $\hat{r}_{c}$ and $\hat{r}_{w}$ respectively denote the rewards corresponding to the correct and incorrect steps in this trajectory. $\zeta$ is a margin hyperparameter, following the previous studies, we set it to 4.
This loss aims to emphasize the ranking of correct versus incorrect trajectories.
It penalizes violations in the relative order between higher-quality and lower-quality steps and focuses on the magnitude of Q-value gaps, thus improving robustness and ranking fidelity in reward modeling.

\subsection{Inference and Evaluation}
During inference, for a given question $q$, we sample $N$ candidate trajectories $D_q = \{\tau_1, \ldots, \tau_N\}$ from the policy model.
We evaluate each trajectory using BiPRM to compute the aggregate score $R(\tau_i)$.
Following the standard Best-of-$N$ protocol, the trajectory with the highest reward is selected as the prediction: $\tau_q^* = \arg\max_{\tau_i \in D_q} R(\tau_i)$.
The accuracy is determined by comparing the final answer extracted from $\tau_q^*$ with the ground truth.
This process rigorously tests the model's ability to identify high-quality solutions from a candidate pool.

\section{Experiments}
\subsection{Experiments Settings}
\paragraph{Datasets.}
We train all models on the Math-Shepherd dataset~\cite{wang2024math}. For evaluation, we employ a two-tiered approach covering both solution-level selection and step-level error detection. 
A detailed summary of the dataset statistics is provided in Table~\ref{tab:datasets}, alongside comprehensive descriptions of data processing in Appendix~\ref{sec:Implementation_Details}.

\textbf{Solution-level Evaluation:} We utilize two widely recognized benchmarks, GSM-Plus~\cite{li2024gsm} and MATH500~\cite{hendrycks2021measuring}. The test set comprises 128 candidate solutions per question, sampled from three diverse policy models (MetaMath-Mistral-7B~\cite{yu2024metamath}, MuggleMath-13B~\cite{li2024query}, Llama-3-70B-Instruct~\cite{grattafiori2024llama}) as provided by \citet{li2024process}.

\textbf{Step-level Evaluation:} We employ ProcessBench~\cite{zheng2025processbench}, a diagnostic benchmark containing 3,400 test cases with expert-annotated error locations. 
This dataset specifically tests the model's precision in identifying the first erroneous step in complex reasoning chains from competition-level problems.

\paragraph{Implementation Details.}
We conduct experiments across three LLM backbones with varying capacities (Rho-Math-1B~\cite{lin2024rho}, Qwen2.5-Math-1.5B~\cite{yang2024qwen2}, Deepseek-Math-7B~\cite{shao2024deepseekmath}) and three objectives (BCE, MSE, Q-value Rankings Loss). 
BiPRM shares identical training configurations with baselines to ensure a rigorous comparison.
Detailed hyperparameter are listed in Appendix~\ref{sec:Implementation_Details}.

\paragraph{Metrics.}
For solution-level evaluation, We report the Best-of-$N$ (BON@$N$) accuracy, which measures the percentage of questions where the top-ranked solution among $N$ candidates is correct.
For step-level evaluation on ProcessBench, we report the F1 score, assessing the model's ability to accurately localize the first error step or correctly identify a flawless solution.

\begin{table*}[!ht]
\setlength{\abovecaptionskip}{6pt}   
    \centering
    \small
    \begin{tabular}{ccc|ccc|ccc}
    \toprule
    \multirow{3}{*}{\shortstack{Sampling\\Policy}} & \multirow{3}{*}{\shortstack{Backbone}} & \multirow{3}{*}{Method} & \multicolumn{3}{c|}{Dataset: MATH500} & \multicolumn{3}{c}{Dataset: GSM-Plus} \\
    &  &  & \multirow{2}{*}{BCE} & \multirow{2}{*}{MSE} & \multirow{2}{*}{\shortstack{Q-value\\rankings}} & \multirow{2}{*}{BCE} & \multirow{2}{*}{MSE} & \multirow{2}{*}{\shortstack{Q-value\\rankings}} \\
    & & & & & & \\
    \toprule
    \multirow{6}{*}{\shortstack{MetaMath-\\Mistral-7B}}
    &  \multirow{2}{*}{Rho-Math-1B} & L2R & 18.48  & 22.32  & 23.36  & 36.64  & 40.09  & 42.06 \\
    &  & Ours & \textbf{23.56} & \textbf{24.96} & \textbf{26.04} & \textbf{45.89} & \textbf{45.28} & \textbf{47.96} \\
    \cmidrule(lr){4-9}
    &  \multirow{2}{*}{Qwen2.5-Math-1.5B} & L2R & 32.80 & 35.68 & 36.20 & 51.90 & 55.87 & 54.81 \\
    &  & Ours & \textbf{38.56} & \textbf{39.80} & \textbf{40.24} & \textbf{55.02} & \textbf{58.11} & \textbf{58.87} \\
    \cmidrule(lr){4-9}
    &  \multirow{2}{*}{Deepseek-Math-7B} & L2R & 31.64  & 34.20  & 32.24  & 53.29  & 56.27  & 54.84 \\
    &  & Ours & \textbf{36.44} & \textbf{37.40} & \textbf{34.48} & \textbf{56.96} & \textbf{59.52} & \textbf{57.13} \\
    \midrule
    \multirow{6}{*}{\shortstack{Muggle-\\Math-13B}} 
    &  \multirow{2}{*}{Rho-Math-1B} & L2R & 18.88  & 16.04  & 18.56  & 33.68  & 38.81  & 41.84  \\
    &  & Ours & \textbf{21.00} & \textbf{19.52} & \textbf{20.56} & \textbf{42.89} & \textbf{44.02} & \textbf{45.09} \\
    \cmidrule(lr){4-9}
    &  \multirow{2}{*}{Qwen2.5-Math-1.5B} & L2R & 25.88 & 31.76 & 30.96 & 53.41 & 53.23 & 54.17 \\
    &  & Ours & \textbf{33.92} & \textbf{34.96} & \textbf{36.08} & \textbf{56.33} & \textbf{57.43} & \textbf{58.46} \\
    \cmidrule(lr){4-9}
    &  \multirow{2}{*}{Deepseek-Math-7B} & L2R & 25.08  & 31.44  & 28.52  & 53.58  & 56.55  & 55.30 \\
    &  & Ours & \textbf{32.64} & \textbf{33.92} & \textbf{30.44} & \textbf{59.57} & \textbf{59.49} & \textbf{57.79} \\
    \midrule
    \multirow{6}{*}{\shortstack{Llama-3-\\70B-Instruct}} 
    &  \multirow{2}{*}{Rho-Math-1B} & L2R & 32.76  & 34.40  & 34.80  & 64.14  & 66.69  & 67.23 \\
    &  & Ours & \textbf{37.32} & \textbf{38.24} & \textbf{39.08} & \textbf{67.73} & \textbf{68.64} & \textbf{68.92}\\
    \cmidrule(lr){4-9}
    &  \multirow{2}{*}{Qwen2.5-Math-1.5B} & L2R & 41.44 & 41.76 & 44.20 & 68.76 & 69.82 & 70.17 \\
    &  & Ours & \textbf{48.48} & \textbf{46.68} & \textbf{47.80} & \textbf{70.44} & \textbf{70.93} & \textbf{71.20} \\
    \cmidrule(lr){4-9}
    &  \multirow{2}{*}{Deepseek-Math-7B} & L2R & 40.24  & 42.52  & 40.56  & 68.88  & 70.07  & 70.75 \\
    &  & Ours & \textbf{43.48} & \textbf{48.12} & \textbf{43.92} & \textbf{71.07} & \textbf{71.21} & \textbf{71.55} \\
    \toprule
    \end{tabular}
    \caption{\textbf{Solution-level} comparison of Best-of-$N$ (BON) performance, averaged from BON@8 to BON@128, across 54 configurations involving two benchmarks, three backbones, three PRM objectives and three sampling policies. BiPRM consistently outperforms the L2R baseline, achieving superior average scores of \textbf{40.37} vs. 36.15 for Rho, \textbf{51.30} vs. 47.38 for Qwen, and \textbf{50.29} vs. 47.00 for Deepseek.
    }
    \label{tab:all results}
\end{table*}

\subsection{Main Results}
\paragraph{Solution-level Verification: Universal Improvements.}
Table~\ref{tab:all results} presents a comprehensive comparison of verification performance across 54 distinct configurations, with detailed per-BON@$n$ results provided in Appendix~\ref{sec:complete_results}.
BiPRM demonstrates robust superiority over the unidirectional L2R baseline across all settings, regardless of model scale or training objective.
Specifically, on the Rho-Math-1B, Qwen2.5-Math-1.5B, and Deepseek-Math-7B backbones, BiPRM achieves average relative improvements of 13.5\%, 10.0\%, and 8.3\%, respectively.
Notably, the performance gain is particularly substantial in challenging scenarios, with a maximum relative improvement of 31.1\% achieved on the Qwen2.5-Math-1.5B backbone.
These consistent gains confirm that the bidirectional evaluation paradigm provides a fundamental modeling advantage.
By integrating future context, BiPRM significantly enhances the ranking precision and stability of the verification process.

\begin{table}[!htbp]
\setlength{\abovecaptionskip}{7pt}   
    \small
    \centering
    \begin{tabular}{cc|ccc}
    \toprule
    \multirow{2}{*}{\shortstack{Subset}} & \multirow{2}{*}{Method} & \multirow{2}{*}{BCE} & \multirow{2}{*}{MSE} & \multirow{2}{*}{\shortstack{Q-value\\rankings}} \\ 
    & & & & \\
    \toprule
    \multirow{2}{*}{GSM8K} & L2R & 37.8 & 50.7 & 43.0 \\
    & Ours & \textbf{46.8} & \textbf{50.8} & \textbf{47.0}  \\
    \midrule
    \multirow{2}{*}{MATH} & L2R & 37.7 & 21.1 & 28.6 \\
    & Ours & \textbf{37.9} & \textbf{23.5} & \textbf{38.0} \\
    \midrule
    \multirow{2}{*}{\shortstack{OlympiadBench}} & L2R & 13.3 & 8.1 & 9.5 \\
    & Ours & \textbf{32.9} & \textbf{12.1} & \textbf{18.7} \\
    \midrule
    \multirow{2}{*}{\shortstack{Omni-MATH}} & L2R & 26.0 & 5.4 & 14.0  \\
    & Ours & \textbf{28.0} & \textbf{7.1} & \textbf{19.9}  \\
    \toprule
    \end{tabular}
    \caption{\textbf{Step-level} error detection performance (F1 Score) using ProcessBench on the Qwen2.5-Math-1.5B backbone. BiPRM consistently outperforms the L2R baseline, with average scores of \textbf{30.2} vs. 24.6.}
    \label{tab:ProcessBench_results}
\end{table}

\paragraph{Step-level Error Detection: ProcessBench Evaluation.}
Beyond selecting the correct final answer, a robust verifier must accurately pinpoint logical flaws within the reasoning process.
We evaluate this fine-grained capability using ProcessBench on the Qwen2.5-Math-1.5B backbone.
As detailed in Table~\ref{tab:ProcessBench_results}, BiPRM consistently outperforms the L2R baseline in F1 scores across diverse subsets.
Quantitatively, our method achieves an average relative improvement of 37.7\% compared to the baseline.
While L2R models often struggle with local inconsistencies, BiPRM leverages retrospective verification to detect errors that are only evident when viewed from a global perspective.
These results validate that our method effectively generalizes to out-of-distribution data and offers superior interpretability by precisely locating logical distinct breaks in reasoning chains.

\begin{table*}[!htbp]
\setlength{\abovecaptionskip}{7pt}
    \small
    \centering
    \setlength{\tabcolsep}{1.4mm}
    \begin{tabular}{l|cccc}
    \toprule
    Model & GSM8K & MATH & OlympiadBench & Omni-MATH \\
    \midrule
    Math-Shepherd-PRM-7B~\cite{wang2024math}  & \underline{47.9} & 29.5 & 24.8 & 23.8 \\
    RLHFlow-PRM-Deepseek-8B~\cite{dong2024rlhf}& 38.8 & 33.8 & 16.9 & 16.9 \\
    Qwen2.5-Math-7B-Math-Shepherd~\cite{zhang2025lessons}& \textbf{62.5} & 31.6 & 13.7 & 7.7 \\
    Qwen2.5-Math-RM-72B~\cite{yang2024qwen2}   & 43.5 & \textbf{47.2} & \textbf{37.6} & \underline{27.4} \\
    \midrule
    Ours (Qwen2.5-Math-1.5B) & 46.8 & \underline{37.9} & \underline{32.9} & \textbf{28.0} \\
    \bottomrule
    \end{tabular}
    \caption{Comparison of F1 scores on ProcessBench against strong external baselines. The best values are highlighted in bold, and the second-best values are underlined. BiPRM maintains excellent stability across difficulty levels despite operating at a smaller parameter scale.}
    \label{tab:external_baselines}
\end{table*}

\paragraph{Comparison with Strong External Baselines.}
To further contextualize the capability of our bidirectional framework, we benchmark BiPRM against several recent high-performance PRM baselines. As illustrated in Table~\ref{tab:external_baselines}, despite utilizing a significantly smaller 1.5B backbone, BiPRM achieves cross-scale superiority, consistently outperforming widely recognized 7B~\cite{wang2024math} and 8B~\cite{dong2024rlhf} models across multiple subsets. Impressively, on the GSM8K and Omni-MATH datasets, our 1.5B BiPRM even surpasses the massive Qwen2.5-Math-RM-72B~\cite{yang2024qwen2}. Furthermore, while the performance of baselines such as Qwen2.5-Math-7B-Math-Shepherd~\cite{zhang2025lessons} degrades drastically as mathematical difficulty increases (dropping from 62.5 on GSM8K to just 7.7 on Omni-MATH), BiPRM maintains remarkable stability. This validates the broad applicability and rigorous structural advantage provided by the bidirectional paradigm.

\subsection{Ablation Studies}
We conduct ablation studies on the Qwen2.5-Math-1.5B backbone under BCE and MSE objectives to quantify component contributions and verify the necessity of the bidirectional architecture.

\paragraph{Impact of Bidirectional Components.}
As shown in Table~\ref{tab:ablation results}, removing either the backward stream (w/o R2L) or the forward stream (w/o L2R) leads to significant performance drops compared to the full BiPRM.
This confirms the inherent complementarity of the two directions.
Furthermore, the dynamic gating mechanism (BiPRM) consistently outperforms the static averaging baseline (w/o Dynamic), particularly on difficult datasets like MATH500 (e.g., 48.48 vs. 44.00 on Llama-3-70B-Instruct policy), validating the effectiveness of context-aware adaptive fusion.

\begin{table}[!ht]
    \centering
    \small
    \setlength{\tabcolsep}{1.3mm}
    \begin{tabular}{cc|cc|cc}
    \toprule
    \multirow{2}{*}{\shortstack{Sampling\\Policy}} & \multirow{2}{*}{\shortstack{Method}} 
    & \multicolumn{2}{c|}{MATH500} 
    & \multicolumn{2}{c}{GSM-Plus} \\
    & & BCE & MSE & BCE & MSE \\
    \midrule
    \multirow{6}{*}{\shortstack{MetaMath-\\Mistral-7B}} 
    & w/o R2L  & 32.80 & 35.68 & 51.90 & 55.87 \\
    & w/o L2R & 32.32 & 32.80 & 52.21 & 51.83 \\
    & w/o Dynamic & 38.16 & 37.44 & 53.96 & 58.02 \\
    & $2\times$L2R & 34.56 & 39.44 & 50.12 & 56.68 \\
    & $2\times$R2L & 33.96 & 36.20 & 52.41 & 53.76 \\
    & BiPRM & \textbf{38.56} & \textbf{39.80} & \textbf{55.02} & \textbf{58.11} \\
    \midrule
    \multirow{6}{*}{\shortstack{Muggle-\\Math-13B}}
    & w/o R2L  & 25.88 & 31.76 & 53.41 & 53.23 \\
    & w/o L2R & 27.60 & 32.00 & 51.12 & 55.92 \\
    & w/o Dynamic & 33.28 & 34.72 & 55.55 & 56.47 \\
    & $2\times$L2R & 30.40 & 32.92 & 50.28 & 54.15 \\
    & $2\times$R2L & 33.16 & 34.32 & 51.65 & 53.65 \\
    & BiPRM & \textbf{33.92} & \textbf{34.96} & \textbf{56.33} & \textbf{57.43} \\
    \midrule
    \multirow{6}{*}{\shortstack{Llama-3-\\70B-Instruct}} 
    & w/o R2L  & 41.44 & 41.76 & 68.76 & 69.82 \\
    & w/o L2R & 40.36 & 39.80 & 69.73 & 69.50 \\
    & w/o Dynamic & 44.00 & 44.88 & 70.07 & 70.57 \\
    & $2\times$L2R & 43.64 & 42.52 & 67.69 & 70.29 \\
    & $2\times$R2L & 41.40 & 44.24 & 68.24 & 69.31 \\
    & BiPRM & \textbf{48.48} & \textbf{46.68} & \textbf{70.44} & \textbf{70.93} \\
    \bottomrule
    \end{tabular}
    \caption{Ablation results on the Qwen2.5-Math-1.5B backbone under BCE and MSE objectives. Metrics report the average BON@$n$ from BON@8 to BON@128. "w/o Dynamic" denotes the static average fusion. "$2 \times$" denotes the ensemble of two independently PRMs trained with different seeds.
    }
    \label{tab:ablation results}
\end{table}

\paragraph{Superiority over Homogeneous Ensembles.}
A potential concern of our bidirectional framework is whether the performance gains stem simply from increased computation (doubled FLOPs) rather than the complementary nature of the bidirectional context.
To address this, we compare BiPRM against homogeneous ensembles that utilize the same computational budget: $2\times$L2R (averaging scores from two independently trained L2R-PRMs) and $2\times$R2L (averaging scores from two R2L-PRMs).
As shown in Table~\ref{tab:ablation results}, BiPRM consistently surpasses these homogeneous ensembles.
For instance, on the MuggleMath/BCE setting, BiPRM (56.33) significantly outperforms both $2\times$L2R (50.28) and $2\times$R2L (51.65).
This demonstrates that the performance leap derives from the synergistic integration of distinct reasoning perspectives rather than mere computational scaling.

\paragraph{Stability in Long-Chain Reasoning.} 
To further investigate the robustness of our dynamic gating mechanism in extended context settings, we conducted targeted experiments isolating long-tail data (sequences strictly $>1024$ tokens). The empirical results demonstrate that dynamic gating successfully preserves robust verification capability even when processing highly extended trajectories, consistently outperforming static averaging methods. For detailed quantitative results and analysis, please refer to Appendix~\ref{sec:appendix_long_chain}.

\begin{figure}[ht]
\setlength{\abovecaptionskip}{7pt}   
    \centering
    \includegraphics[width=1\linewidth]{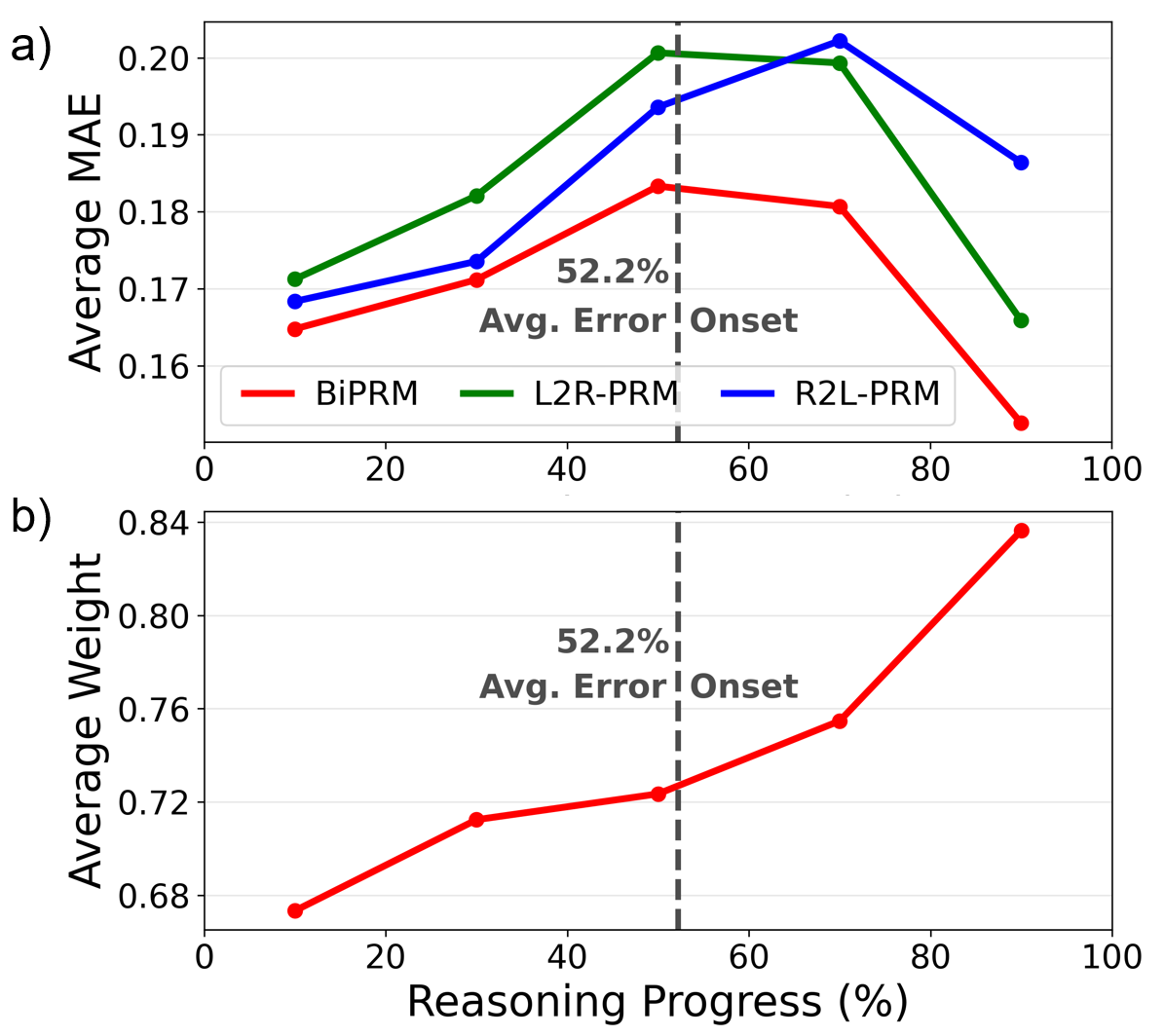}
    \caption{Analysis of (a) step-wise MAE error distribution and (b) the evolution of dynamic gating weights on the Math-Shepherd test set. The horizontal axis represents the normalized progress of reasoning steps within a solution trajectory. }
    \label{fig:mae_sigma_step}
\end{figure}

\subsection{Analysis}
\label{sec:analysis}
In this section, we analyze the characteristics of different reward modeling paradigms through quantitative error trends, the evolution of learned gating weights, and practical inference efficiency.

\paragraph{Error Distribution across Reasoning Steps.} Figure~\ref{fig:mae_sigma_step}(a) illustrates the MAE trajectories relative to the normalized reasoning progress. 
The unidirectional baselines exhibit distinct position-dependent biases. The vertical dashed line marks the Average Error Onset at 52.2\%, which serves as the critical boundary shifting from correct premises to subsequent logical failures. 
We observe that all models encounter their highest prediction uncertainty in this transition zone, resulting in peak MAE values around the center of the trajectory. 
Nevertheless, the unidirectional baselines exhibit divergent behaviors away from this boundary. 
The L2R-PRM (green) shows reduced errors towards the final steps (90\%) as it benefits from accumulated forward context. 
Conversely, the R2L-PRM (blue) achieves lower error rates in the early stages (10\% to 30\%) because the reverse stream effectively treats these initial steps as terminal stages with full context visibility.
Crucially, BiPRM (red) consistently maintains the lowest MAE throughout the entire process, particularly effectively suppressing the error spike in the high-difficulty transition region. 
This suggests that BiPRM successfully fuses these complementary strengths to eliminate localized blind spots.

\paragraph{Evolution of Dynamic Gating Weights.}
To further validate the adaptive nature of our fusion mechanism, we visualize the average value of the learnable gating weight $\sigma_t$ across the reasoning progress in Figure~\ref{fig:mae_sigma_step}(b).
Recall that $\sigma_t$ represents the weighting coefficient for the L2R stream (i.e., $r_t^{\text{BiPRM}} = \sigma_t \cdot r_t^{\text{L2R}} + (1 - \sigma_t) \cdot r_t^{\text{R2L}}$).
As observed in the figure, $\sigma_t$ exhibits a monotonically increasing trend as reasoning progresses.
In the early stages (0\%-30\%), $\sigma_t$ remains relatively low, indicating that the model automatically learns to rely more heavily on the R2L stream ($1-\sigma_t$ is high).
This aligns with our findings in the error distribution analysis that the backward view provides superior verification for initial steps.
Conversely, as the reasoning approaches the final answer (70\%-100\%), $\sigma_t$ increases significantly, shifting the focus towards the L2R stream to leverage the accumulated forward context.
This adaptive weight allocation confirms that BiPRM effectively captures the complementary strengths of the dual streams, dynamically prioritizing the most informative context for each specific step.

\begin{table}[!htbp]
    \small
    \centering
    \setlength{\tabcolsep}{1.3mm}
    \begin{tabular}{cc|ccc}
    \toprule
    \multirow{2}{*}{Subset} & \multirow{2}{*}{Method} & \multicolumn{1}{c}{\multirow{2}{*}{BCE}} & \multicolumn{1}{c}{\multirow{2}{*}{MSE}} & \multicolumn{1}{c}{\multirow{2}{*}{\shortstack{Q-value\\rankings}}} \\
    & & & & \\
    \midrule
    \multirow{2}{*}{GSM8K} & L2R & 35.9 ({$\downarrow$1.9}) & 44.7 ({$\downarrow$6.0}) & 38.5 ({$\downarrow$4.5}) \\
    & Ours & \textbf{43.9} ({$\downarrow$2.9}) & \textbf{45.6} ({$\downarrow$5.2}) & \textbf{44.3} ({$\downarrow$2.7}) \\
    \midrule
    \multirow{2}{*}{MATH} & L2R & 34.9 ({$\downarrow$2.8}) & 19.8 ({$\downarrow$1.3}) & 27.4 ({$\downarrow$1.2}) \\
    & Ours & \textbf{36.6} ({$\downarrow$1.3}) & \textbf{21.7} ({$\downarrow$1.8}) & \textbf{37.2} ({$\downarrow$0.8}) \\
    \midrule
    \multirow{2}{*}{\shortstack{Olympiad\\Bench}} & L2R & 13.1 ({$\downarrow$0.2}) & 6.4 ({$\downarrow$1.7}) & 9.2 ({$\downarrow$0.3}) \\
    & Ours & \textbf{31.1} ({$\downarrow$1.8}) & \textbf{10.5} ({$\downarrow$1.6}) & \textbf{18.7} ({$\downarrow$0.0}) \\
    \midrule
    \multirow{2}{*}{\shortstack{Omni-\\MATH}} & L2R & 24.5 ({$\downarrow$1.5}) & 4.6 ({$\downarrow$0.8}) & 14.0 ({$\downarrow$0.0}) \\
    & Ours & \textbf{27.6} ({$\downarrow$0.4}) & \textbf{6.8} ({$\downarrow$0.3}) & \textbf{19.7} ({$\downarrow$0.2}) \\
    \bottomrule
    \end{tabular}
    \caption{Sensitivity analysis under step boundary perturbations on the Qwen2.5-Math-1.5B backbone. Values represent the Perturbed F1 Score and the absolute degradation ($\downarrow$) relative to the original clean results in Table~\ref{tab:ProcessBench_results}. BiPRM exhibits notably smaller performance drops.}
    \label{tab:noise_robustness}
\end{table}

\paragraph{Robustness to Ambiguous Step Boundaries.}
To address potential practical concerns regarding noisy and ambiguous step boundaries, we conducted a targeted sensitivity analysis by artificially introducing boundary perturbations (merging, splitting, and noise insertion) into the ProcessBench dataset. As shown in Table~\ref{tab:noise_robustness}, while all models naturally exhibit performance degradation under these adverse conditions, BiPRM demonstrates significantly superior stability. On average, BiPRM experiences an F1 drop of only 5.24\%, compared to a 7.52\% drop for the L2R baseline. This confirms that the bidirectional architecture fundamentally enhances structural robustness, making it highly reliable in real-world scenarios containing inconsistent step segmentations.

\begin{table*}[!ht]
\setlength{\abovecaptionskip}{8pt}
    \setlength{\tabcolsep}{1.5mm}
    \begin{tabularx}{\textwidth}{X|cccc}
    \toprule
    \textbf{Question:} Given a sequence $\{a_n\}$ defined by $a_1 = 1$ and $a_{n+1} = 2a_n + 1$, find the general term formula. & \multirow{2}{*}{Label} & \multirow{2}{*}{\shortstack{Bi-\\PRM}} & \multirow{2}{*}{\shortstack{L2R\\-PRM}} & \multirow{2}{*}{\shortstack{R2L\\-PRM}} \\
    \midrule
    \textbf{Step 1:} Compute the first few terms: $a_2 = 2a_1 + 1 = 3$, $a_3 = 2a_2 + 1 = 7$, $a_4 = 2a_3 + 1 = 15, \ldots$. & 1 & 1.00 & 1.00 & 1.00 \\
    \textbf{\textcolor{deepred}{Step 2:}} Observing that $a_n + 1$ yields powers of 2, we conjecture: \textcolor{deepred}{$\bm{a_n = 2n - 1}$}. & 0 & 0.31 & 1.00 & 0.29 \\
    \textbf{Step 3:} Base case $(n=1)$: $a_1 = 1 = 2 \times 1 - 1$. & 0 & 0.38 & 1.00 & 0.38 \\
    \textbf{\textcolor{deepred}{Step 4:}} Assume $a_k = 2k - 1$ holds for $n = k$, then for $n = k + 1$, $a_{k+1} = 2a_k + 1 = \textbf{\textcolor{deepred}{2(k + 1) - 1}}$. Hence, it holds for $n = k + 1$.  & \raisebox{-13pt}{0} & \raisebox{-13pt}{0.01} & \raisebox{-13pt}{1.00} & \raisebox{-13pt}{0.00} \\
    \textbf{Step 5:} By the principle of mathematical induction, the formula is valid for all natural numbers $n$. The general term formula is $a_n = 2n - 1$. & \raisebox{-13pt}{0} & \raisebox{-13pt}{0.17} & \raisebox{-13pt}{0.99} & \raisebox{-13pt}{0.08} \\
    \bottomrule
    \end{tabularx}
    \caption{A representative case study comparing step-wise reward scores of BiPRM, L2R-PRM, and R2L-PRM. The critical error stems from an incorrect conjecture in \textbf{Step 2} (highlighted in red), which the model attempts to justify through a forced derivation in \textbf{Step 4}. While L2R-PRM is deceived by the superficial coherence, R2L-PRM successfully identifies the logical fallacy through retrospective verification.}
    \label{tab:case study}
\end{table*}

\paragraph{Inference Latency.} 
\label{sec:Latency}
A primary concern regarding bidirectional architectures is the potential doubling of inference time. 
However, since the L2R and R2L evaluation streams operate independently prior to the final gating fusion, they can be processed concurrently. 
By stacking the forward and reverse inputs into a single batch, we leverage the parallel computing capabilities of the GPU to minimize latency. 
As shown in Table~\ref{tab:efficiency_comparison}, empirical results demonstrate that the average wall-clock time for scoring a single solution increases only marginally from 27.982 ms for the L2R baseline to 29.393 ms for BiPRM.
This corresponds to a relative latency overhead of approximately 5\%, demonstrating that our parallel implementation effectively mitigates the temporal cost of dual-stream evaluation, making BiPRM highly practical for real-world deployment.
Further discussions on computational efficiency, including detailed comparisons of model parameters, measured TFLOPS, and peak memory usage, are provided in Appendix~\ref{sec:appendix_efficiency}.

\begin{table}[!htbp]
    \centering
    \small
    \setlength{\tabcolsep}{1.8mm}
    \begin{tabular}{c|ccc}
    \toprule
    Model & Params & TFLOPS & Wall-clock Time$^\dagger$ \\
    \midrule
    L2R-PRM & 1.543B & 10.034 & 27.982 ms  \\
    BiPRM (Ours) & 1.548B & 19.106 & 29.393 ms  \\
    \bottomrule
    \end{tabular}
    \caption{Quantitative comparison of model parameters, measured computational cost (TFLOPS), and inference time latency between L2R-PRM and BiPRM on the Qwen2.5-Math-1.5B backbone. $\dagger$: Measured as the average inference time per single solution.}
    \label{tab:efficiency_comparison}
\end{table}

\paragraph{Case Study.}
Table~\ref{tab:case study} presents a representative example where the model constructs a superficially coherent proof based on an incorrect hypothesis in Step 2.
The L2R baseline fails to detect this error because it views the hypothesis in Step 2 as a plausible conjecture pending verification, assigning it a high score based on local coherence.
In contrast, the R2L stream correctly penalizes the final conclusion in Step 5 without needing to trace the intermediate derivation.
By evaluating the final answer directly against the problem statement, the R2L stream identifies the fundamental mathematical contradiction between the derived formula and the sequence definition, thereby effectively filtering out the erroneous trajectory.
For a more comprehensive qualitative analysis on success and failure cases, please refer to Appendix~\ref{sec:appendix_cases}.

\section{Conclusion}
In this paper, we proposed BiPRM, a novel bidirectional evaluation paradigm designed to address the limitations of unidirectional PRMs in accessing global context.
By synergizing a parallel reverse evaluation stream with a dynamic gating mechanism, BiPRM enables effective retrospective verification while incurring negligible computational overhead.
Extensive experiments demonstrate that our method consistently outperforms L2R baselines in both solution-level ranking and step-level error localization.
These findings underscore the necessity of bidirectional context for robust reasoning supervision and offer a promising direction for future research in process reward modeling.

\section*{Limitations}
While the promising results achieved by BiPRM, certain limitations remain.
\paragraph{Computational Cost.} The bidirectional evaluation mechanism inherently necessitates processing the input sequence twice, leading to a two-fold increase in theoretical floating-point operations (FLOPs) compared to unidirectional baselines. This increased energy consumption is an unavoidable trade-off for the enhanced verification capability. However, from the perspective of practical deployment, we effectively mitigate the impact on user experience through a parallel execution strategy. As demonstrated in our efficiency analysis (Section~\ref{sec:Latency}), this implementation ensures that the actual inference time latency increases by merely 5\%, maintaining the method's feasibility for real-time applications.

\paragraph{Generalization across Domains.} Our experimental validation is currently confined to the domain of mathematical reasoning. While mathematics serves as a rigorous testbed for evaluating logical consistency and stepwise correctness, the generalizability of the bidirectional verification paradigm to other complex tasks remains to be verified. We leave the exploration of broader application scenarios, such as code generation, open-ended commonsense reasoning, and symbolic logic tasks, to future research. We hope that our findings will serve as a foundation for developing more robust and generalized process supervision frameworks in these diverse fields.

\section*{Acknowledgments}
This work was funded by the National Natural Science Foundation of China (No. 62106165), the Jiangsu Major Science and Technology Special Fund for Innovative Biologics (No. BG2025062), and the Priority Academic Program Development of Jiangsu Higher Education Institutions.

\bibliography{custom}
\appendix
\clearpage

\section{Implementation Details}
\label{sec:Implementation_Details}
\paragraph{Data Preparation.} 
The training corpus is derived from the Math-Shepherd dataset, which comprises mathematical questions paired with multi-step reasoning solutions and step-level supervision signals. 
To ensure data quality, we filter out instances containing only a single reasoning step and partition the remaining data into training and validation subsets using a 95:5 ratio. 
For evaluation, the test set includes 128 candidate solutions generated by each policy model for every question in the benchmarks, resulting in six distinct evaluation sets.

\paragraph{Training Configuration.} 
We independently train both the unidirectional PRMs and our proposed BiPRMs across all nine combinations formed by the LLM backbones and PRM objectives. 
Crucially, each pair of PRM and BiPRM models shares identical training configurations to guarantee a fair comparison. 
All models are trained using 8 NVIDIA RTX A5000 GPUs. 
Regarding the software environment, we use the following package versions: \texttt{torch==2.3.1+cu118, trl==0.8.0, transformers==4.43.0, accelerate==0.33.0, deepspeed==0.13.1, nvidia-nccl-cu12==2.20.5}. 
We employ the ZeRO-3 optimization stage of DeepSpeed with bfloat16 precision. 
Gradient checkpointing is set to true for Deepseek-Math-7B, while it is disabled for both Qwen2.5-Math-1.5B and Rho-Math-1B. 
The specific hyperparameters for all experiments are detailed in Table~\ref{tab:hyperparameters}.

\begin{table}[!h]
    \centering
    \begin{tabular}{c|c}
    \toprule
    Hyperparameters & Value \\
    \midrule
    epoch & 1 \\
    learnning rate & 3e-5 \\
    optimizer & AdamW \\
    scheduler & linear \\
    seed & 1106 \\
    batch size per GPU & 2 \\
    gradient accumulation steps & 4 \\
    \bottomrule
    \end{tabular}
    \caption{Hyperparameters in all experimental configurations.}
    \label{tab:hyperparameters}
\end{table}

\section{Computational Efficiency Analysis}
\label{sec:appendix_efficiency}
In this section, we provide a supplementary analysis of the computational efficiency between BiPRM and the standard unidirectional L2R-PRM.
All measurements were conducted on the Qwen2.5-Math-1.5B backbone using a single NVIDIA RTX A5000 GPU to ensure a controlled experimental environment.
Our evaluation focuses on four key dimensions: model parameter size, measured computational cost (TFLOPS), peak memory usage, and the actual inference latency (wall-clock time).

As shown in Table~\ref{tab:efficiency_comparison}, the parameter overhead introduced by our dynamic gating module is a negligible 0.3\%, increasing the total count from 1.543B to 1.548B. In terms of measured computational cost, BiPRM inherently requires processing the input sequence twice, resulting in an empirical TFLOPS increase from 10.034 to 19.106. 
Crucially, despite the doubling of empirical compute, BiPRM achieves a highly favorable and practical trade-off for real-world deployment. Parallel execution of the two streams causes only a marginal relative increase in overall peak memory usage, rising from 6.43 GB to 6.65 GB. Consequently, the average wall-clock time for scoring a single solution increases by just 5\%, from 27.982 ms for the L2R baseline to 29.393 ms for BiPRM. This demonstrates that our implementation effectively exchanges an acceptable, low-impact overhead in computation and memory for substantially enhanced verification capabilities  with minimal impact on user-perceived latency.

\section{Complete experimental results}
\label{sec:complete_results}
We present the complete experimental results of Rho-Math-1B, Qwen2.5-Math-1.5B and Deepseek-Math-7B in Tables~\ref{tab:rho results}, ~\ref{tab:Qwen results} and~\ref{tab:deepseek results}. Including specific values of bon@8 to bon@128 across the two benchmarks and three sampling policies.

\section{Stability of Dynamic Gating in Long-Chain Scenarios}
\label{sec:appendix_long_chain}
To thoroughly analyze the stability of our dynamic gating mechanism in extremely long-chain reasoning scenarios, we conducted additional evaluations focusing on long-tail data with sequence lengths strictly greater than 1024 tokens. This evaluation was performed using the Qwen2.5-Math-1.5B backbone under the BCE objective.

As detailed in Table~\ref{tab:long_chain_stability}, the average sequence lengths in these long-tail subsets significantly increase compared to the original datasets (e.g., from 988 to 1457 on MATH500). While all models inevitably experience a certain degree of performance degradation when handling such exceptionally long reasoning chains, the "Dynamic Gating" strategy consistently demonstrates superior resilience compared to the "Static Averaging" baseline. Overall, static averaging exhibits an average performance drop of 37.56\% relative to its standard results, whereas our dynamic gating mechanism limits this drop to 31.40\%. This empirical evidence validates that dynamic context-aware fusion is crucial for maintaining stable verification performance on extended contexts.

\begin{table}[!ht]
    \centering
    \small
    \setlength{\tabcolsep}{1.2mm}
    \begin{tabular}{cc|cc}
    \toprule
    \multirow{2}{*}{\shortstack{Sampling\\Policy}} & \multirow{2}{*}{Method} & MATH500 & GSM-Plus \\
    & & \scriptsize{[988 $\rightarrow$ 1457]} & \scriptsize{[779 $\rightarrow$ 1201]} \\
    \midrule
    \multirow{2}{*}{\shortstack{MetaMath-\\Mistral-7B}} & Static Avg. & 38.16 $\rightarrow$ 16.52 & 53.96 $\rightarrow$ 40.22 \\
    & Dynamic & \textbf{38.56 $\rightarrow$ 19.79} & \textbf{55.02 $\rightarrow$ 43.18} \\
    \midrule
    \multirow{2}{*}{\shortstack{Muggle-\\Math-13B}} & Static Avg. & 33.28 $\rightarrow$ 9.41 & 55.55 $\rightarrow$ 41.83 \\
    & Dynamic & \textbf{33.92 $\rightarrow$ 11.81} & \textbf{56.33 $\rightarrow$ 45.26} \\
    \midrule
    \multirow{2}{*}{\shortstack{Llama-3-\\70B-Instruct}} & Static Avg. & 44.00 $\rightarrow$ 23.02 & 70.07 $\rightarrow$ 53.22 \\
    & Dynamic & \textbf{48.48 $\rightarrow$ 29.11} & \textbf{70.44 $\rightarrow$ 58.54} \\
    \bottomrule
    \end{tabular}
    \caption{Performance (Average BON@$n$) comparison between Static Averaging and Dynamic Gating on original vs. long-tail data (sequences $>1024$ tokens). Bracketed values indicate the average sequence lengths of the original versus the long-tail subset.}
    \label{tab:long_chain_stability}
\end{table}

\begin{table*}[!ht]
    \centering
    \setlength{\tabcolsep}{1.9mm}
    \small
    \begin{tabular}{ccc|ccccc|ccccc}
    \toprule
    \multirow{2}{*}{\shortstack{Sampling\\Policy}} & \multirow{2}{*}{\shortstack{PRM\\Objective}} & \multirow{2}{*}{Method} & \multicolumn{5}{c|}{Dataset: MATH500} & \multicolumn{5}{c}{Dataset: GSM-Plus} \\
    &  &  & @8 & @16 & @32 & @64 & @128 & @8 & @16 & @32 & @64 & @128 \\
    \midrule
    \multirow{6}{*}{\shortstack{MetaMath-\\Mistral-7B}}
    &  \multirow{2}{*}{BCE} & L2R & 21.20  & 20.80  & 17.40  & 15.80  & 17.20  & 43.62  & 40.21  & 36.79  & 33.04  & 29.54 \\
    &  & Ours & \textbf{24.20} & \textbf{24.00} & \textbf{24.60} & \textbf{22.60} & \textbf{22.40} & \textbf{47.46} & \textbf{46.71} & \textbf{45.79} & \textbf{44.92} & \textbf{44.58} \\
    \cmidrule(lr){2-13}
    &  \multirow{2}{*}{MSE} & L2R & 24.80  & 23.60  & 22.80  & 21.60  & 18.80  & 47.96  & 44.50  & 41.54  & 36.67  & 29.79 \\
    &  & Ours & \textbf{27.00} & \textbf{25.80} & \textbf{25.20} & \textbf{23.40} & \textbf{23.40} & \textbf{50.08} & \textbf{47.50} & \textbf{45.46} & \textbf{43.12} & \textbf{40.25} \\
    \cmidrule(lr){2-13}
    &  \multirow{2}{*}{\shortstack{Q-value\\rankings}} & L2R & 25.60  & 24.20  & 23.80  & 22.60  & 20.60  & 49.71  & 46.54  & 42.88  & 38.25  & 32.92 \\
    &  & Ours & \textbf{26.80} & \textbf{26.40} & \textbf{26.40} & \textbf{26.80} & \textbf{23.80} & \textbf{51.54} & \textbf{49.79} & \textbf{47.88} & \textbf{46.00} & \textbf{44.58} \\
    \midrule\midrule
    \multirow{6}{*}{\shortstack{Muggle-\\Math-13B}} 
    &  \multirow{2}{*}{BCE} & L2R & 20.00  & 21.00  & 19.00  & 18.60  & 15.80  & 41.38  & 36.67  & 34.12  & 30.21  & 26.04 \\
    &  & Ours & \textbf{23.80} & \textbf{21.80} & \textbf{19.20} & \textbf{20.60} & \textbf{19.60} & \textbf{43.83} & \textbf{43.58} & \textbf{43.12} & \textbf{42.00} & \textbf{41.92} \\
    \cmidrule(lr){2-13}
    &  \multirow{2}{*}{MSE} & L2R & 18.40  & 17.60  & 16.20  & 14.60  & 13.40  & 45.50  & 43.46  & 39.29  & 34.25  & 31.54 \\
    &  & Ours & \textbf{21.60} & \textbf{21.20} & \textbf{19.20} & \textbf{18.80} & \textbf{16.80} & \textbf{47.79} & \textbf{47.00} & \textbf{44.71} & \textbf{41.88} & \textbf{38.71} \\
    \cmidrule(lr){2-13}
    &  \multirow{2}{*}{\shortstack{Q-value\\rankings}} & L2R & 20.20  & 17.80  & 18.40  & 19.20  & 17.20  & 46.92  & 45.08  & 42.46  & 38.71  & 36.04 \\
    &  & Ours & \textbf{21.00} & \textbf{21.60} & \textbf{21.00} & \textbf{20.80} & \textbf{18.40} & \textbf{47.88} & \textbf{46.88} & \textbf{45.08} & \textbf{43.54} & \textbf{42.08} \\
    \midrule\midrule
    \multirow{6}{*}{\shortstack{Llama-3-\\70B-Instruct}} 
    &  \multirow{2}{*}{BCE} & L2R & \textbf{38.60}  & 36.20  & 33.40  & 28.60  & 27.00  & 66.67  & 65.92  & 64.42  & 62.88  & 60.83 \\
    &  & Ours & 37.20 & \textbf{37.80} & \textbf{37.40} & \textbf{37.40} & \textbf{36.80} & \textbf{67.88} & \textbf{68.21} & \textbf{67.83} & \textbf{67.67} & \textbf{67.08} \\
    \cmidrule(lr){2-13}
    &  \multirow{2}{*}{MSE} & L2R & 38.20  & 36.00  & 35.60  & 32.80  & 29.40  & 69.08  & 68.21  & 67.54  & 66.04  & 62.58 \\
    &  & Ours & \textbf{41.00} & \textbf{37.80} & \textbf{37.80} & \textbf{37.60} & \textbf{37.00} & \textbf{70.54} & \textbf{69.96} & \textbf{68.88} & \textbf{67.46} & \textbf{66.38} \\
    \cmidrule(lr){2-13}
    &  \multirow{2}{*}{\shortstack{Q-value\\rankings}} & L2R & 39.00  & 35.40  & 34.60  & 33.20  & 31.80  & 69.83  & 68.46  & 67.17  & 66.29  & 64.38 \\
    &  & Ours & \textbf{41.00} & \textbf{39.60} & \textbf{39.20} & \textbf{38.20} & \textbf{37.40} & \textbf{71.46} & \textbf{69.54} & \textbf{68.96} & \textbf{68.25} & \textbf{66.38} \\
    \bottomrule
    \end{tabular}
    \caption{\textbf{Rho-Math-1B} results measured by Best-of-N (BON@$n$) accuracy across two benchmarks, three PRM objectives and three sampling policies.}
    \label{tab:rho results}
\end{table*}

\section{Additional Qualitative Analysis}
\label{sec:appendix_cases}
In this section, we provide a deeper analysis of BiPRM's performance across different error types.
We verify its robustness in detecting calculation errors and discuss specific scenarios where the R2L stream may face challenges.

\paragraph{Effectiveness in Calculation Error Detection.}
Table~\ref{tab:case-good} presents a case involving a quadratic function.
In this instance, a critical calculation error occurs at Step 3 regarding the vertex coordinate.
Consistent with our error distribution analysis, the R2L-PRM precisely identifies this early-stage mistake with a score of $0.05$, whereas the L2R-PRM fails to penalize it immediately, assigning an ambiguous score of $0.46$.
BiPRM effectively integrates these divergent signals to provide a precise and robust evaluation, demonstrating that the bidirectional mechanism is beneficial not only for logical reasoning but also for rigorous numerical verification.

\paragraph{Limitations and Failure Analysis.}
While BiPRM shows significant improvements in most scenarios, it exhibits certain limitations in specific arithmetic inconsistencies.
Table~\ref{tab:case-bad} illustrates a failure case involving a circle geometry problem.
The error arises in Step 4, where the model omits the denominator "2" during algebraic simplification ($\pi r^2 / (2\pi r) = 20 \rightarrow r^2 / r = 20$).
Although Step 4 is mathematically incorrect given Step 3, the transition from Step 4 to Step 5 ($r=20$) is internally consistent.
Since the R2L mechanism primarily evaluates whether the current step can logically lead to the subsequent steps (retrospective consistency), it assigns a relatively high score ($0.63$) to Step 4 because the derivation from Step 4 to the end is smooth.
This suggests that R2L is more sensitive to global logical coherence than to local isolated arithmetic skips.
However, note that the L2R stream successfully identifies this error ($0.01$), and BiPRM's gating mechanism partially mitigates the R2L failure, resulting in a final score of $0.18$.
This case highlights the necessity of fusing both directions to handle disjointed arithmetic errors effectively.

\begin{table*}[!ht]
    \centering
    \setlength{\tabcolsep}{1.9mm}
    \small
    \begin{tabular}{ccc|ccccc|ccccc}
    \toprule
    \multirow{2}{*}{\shortstack{Sampling\\Policy}} & \multirow{2}{*}{\shortstack{PRM\\Objective}} & \multirow{2}{*}{Method} & \multicolumn{5}{c|}{Dataset: MATH500} & \multicolumn{5}{c}{Dataset: GSM-Plus} \\
    &  &  & @8 & @16 & @32 & @64 & @128 & @8 & @16 & @32 & @64 & @128   \\
    \toprule
    \multirow{6}{*}{\shortstack{MetaMath-\\Mistral-7B}}
    &  \multirow{2}{*}{BCE} & L2R & 31.60 & 32.80 & 33.40 & 34.00 & 32.20  & 53.92 & 53.96 & 52.62 & 51.21 & 47.79  \\
    &  & Ours & \textbf{34.20} & \textbf{38.60} & \textbf{38.80} & \textbf{40.60} & \textbf{40.60} & \textbf{56.46} & \textbf{56.42} & \textbf{55.75} & \textbf{54.46} & \textbf{52.00} \\
    \cmidrule(lr){2-13}
    &  \multirow{2}{*}{MSE} & L2R & 32.40 & 36.20 & 35.80 & 36.40 & 37.60 & 57.33 & 56.50 & 55.96 & 55.75 & 53.83  \\
    &  & Ours & \textbf{36.20} & \textbf{38.20} & \textbf{40.20} & \textbf{42.20} & \textbf{42.20} & \textbf{58.33} & \textbf{57.71} & \textbf{58.25} & \textbf{58.58} & \textbf{57.67} \\
    \cmidrule(lr){2-13}
    &  \multirow{2}{*}{\shortstack{Q-value\\rankings}} & L2R & 32.40 & 35.40 & 36.80 & 37.80 & 38.60 & 54.75 & 54.88 & 54.88 & 54.58 & 54.96  \\
    &  & Ours & \textbf{36.00} & \textbf{39.00} & \textbf{42.00} & \textbf{42.60} & \textbf{41.60} & \textbf{57.92} & \textbf{58.33} & \textbf{59.38} & \textbf{59.71} & \textbf{59.00} \\
    \midrule\midrule
    \multirow{6}{*}{\shortstack{Muggle-\\Math-13B}} 
    &  \multirow{2}{*}{BCE} & L2R & 26.20 & 26.00 & 26.80 & 26.20 & 24.20 & 52.79 & 54.12 & 53.46 & 53.83 & 52.83  \\
    &  & Ours & \textbf{30.60} & \textbf{34.00} & \textbf{35.60} & \textbf{36.20} & \textbf{33.20} & \textbf{56.04} & \textbf{57.54} & \textbf{56.62} & \textbf{55.92} & \textbf{55.54} \\
    \cmidrule(lr){2-13}
    &  \multirow{2}{*}{MSE} & L2R & 28.60 & 32.00 & 33.00 & 32.80 & 32.40 & 54.08 & 54.12 & 53.29 & 52.92 & 51.75  \\
    &  & Ours & \textbf{32.60} & \textbf{34.00} & \textbf{34.80} & \textbf{36.60} & \textbf{36.80} & \textbf{56.29} & \textbf{57.96} & \textbf{57.79} & \textbf{58.08} & \textbf{57.04} \\
    \cmidrule(lr){2-13}
    &  \multirow{2}{*}{\shortstack{Q-value\\rankings}} & L2R & 27.20 & 32.00 & 31.80 & 32.40 & 31.40 & 53.42 & 55.04 & 54.04 & 54.00 & 54.33 \\
    &  & Ours & \textbf{33.40} & \textbf{35.20} & \textbf{36.60} & \textbf{37.40} & \textbf{37.80} & \textbf{57.75} & \textbf{58.75} & \textbf{59.04} & \textbf{58.50} & \textbf{58.25} \\
    \midrule\midrule
    \multirow{6}{*}{\shortstack{Llama-3-\\70B-Instruct}} 
    &  \multirow{2}{*}{BCE} & L2R & 42.40 & 42.80 & 41.20 & 40.80 & 40.00 & 70.04 & 69.71 & 70.00 & 67.58 & 66.46 \\
    &  & Ours & \textbf{47.40} & \textbf{47.60} & \textbf{46.40} & \textbf{50.80} & \textbf{50.20} & \textbf{72.42} & \textbf{71.46} & \textbf{70.83} & \textbf{69.33} & \textbf{68.17} \\
    \cmidrule(lr){2-13}
    &  \multirow{2}{*}{MSE} & L2R & 44.60 & 43.40 & 42.00 & 40.40 & 38.40 & 70.96 & 70.92 & 70.29 & 68.79 & 68.12  \\
    &  & Ours & \textbf{45.20} & \textbf{46.80} & \textbf{45.80} & \textbf{48.60} & \textbf{47.00} & \textbf{72.33} & \textbf{72.17} & \textbf{70.88} & \textbf{70.50} & \textbf{68.79} \\
    \cmidrule(lr){2-13}
    &  \multirow{2}{*}{\shortstack{Q-value\\rankings}} & L2R & 44.20 & 44.40 & 45.40 & 44.00 & 43.00 & 71.21 & 71.12 & 70.21 & 69.29 & 69.00  \\
    &  & Ours & \textbf{46.80} & \textbf{47.40} & \textbf{47.20} & \textbf{50.20} & \textbf{47.40} & \textbf{71.58} & \textbf{71.46} & \textbf{71.79} & \textbf{70.62} & \textbf{70.54} \\
    \toprule
    \end{tabular}
    \caption{\textbf{Qwen2.5-Math-1.5B} results measured by Best-of-N (BON@$n$) accuracy across two benchmarks, three PRM objectives and three sampling policies.
    }
    \label{tab:Qwen results}
\end{table*}

\begin{table*}[!ht]
    \centering
    \small
    \setlength{\tabcolsep}{1.9mm}
    \begin{tabular}{ccc|ccccc|ccccc}
    \toprule
    \multirow{2}{*}{\shortstack{Sampling\\Policy}} & \multirow{2}{*}{\shortstack{PRM\\Objective}} & \multirow{2}{*}{Method} & \multicolumn{5}{c|}{Dataset: MATH500} & \multicolumn{5}{c}{Dataset: GSM-Plus} \\
    &  &  & @8 & @16 & @32 & @64 & @128 & @8 & @16 & @32 & @64 & @128 \\
    \midrule
    \multirow{6}{*}{\shortstack{MetaMath-\\Mistral-7B}}
    &  \multirow{2}{*}{BCE} & L2R & 29.80  & 33.00  & 31.60  & 32.20  & 31.60  & 55.50  & 54.88  & 54.38  & 52.21  & 49.50 \\
    &  & Ours & \textbf{35.00} & \textbf{36.20} & \textbf{36.20} & \textbf{37.00} & \textbf{37.80} & \textbf{57.08} & \textbf{58.00} & \textbf{57.42} & \textbf{56.50} & \textbf{55.79} \\
    \cmidrule(lr){2-13}
    &  \multirow{2}{*}{MSE} & L2R & 32.20  & 34.80  & 35.40  & 35.00  & 33.60  & 57.75  & 57.04  & 56.67  & 55.17  & 54.71 \\
    &  & Ours & \textbf{33.60} & \textbf{36.60} & \textbf{39.20} & \textbf{40.00} & \textbf{37.60} & \textbf{58.88} & \textbf{60.29} & \textbf{59.58} & \textbf{59.46} & \textbf{59.38} \\
    \cmidrule(lr){2-13}
    &  \multirow{2}{*}{\shortstack{Q-value\\rankings}} & L2R & 31.60  & 32.40  & 32.40  & 33.00  & 31.80  & 55.29  & 54.92  & 54.88  & 55.00  & 54.12 \\
    &  & Ours & \textbf{33.60} & \textbf{35.00} & \textbf{34.60} & \textbf{36.00} & \textbf{33.20} & \textbf{57.08} & \textbf{57.54} & \textbf{57.71} & \textbf{57.04} & \textbf{56.29} \\
    \midrule\midrule
    \multirow{6}{*}{\shortstack{Muggle-\\Math-13B}} 
    &  \multirow{2}{*}{BCE} & L2R & 25.40  & 24.20  & 25.00  & 24.80  & 26.00  & 54.17  & 55.08  & 53.46  & 53.21  & 51.96 \\
    &  & Ours & \textbf{27.60} & \textbf{32.00} & \textbf{33.80} & \textbf{35.60} & \textbf{34.20} & \textbf{58.33} & \textbf{59.33} & \textbf{60.58} & \textbf{59.71} & \textbf{59.88} \\
    \cmidrule(lr){2-13}
    &  \multirow{2}{*}{MSE} & L2R & 29.20  & 31.80  & 32.40  & 32.00  & 31.80  & 56.83  & 57.83  & 56.58  & 56.46  & 55.04 \\
    &  & Ours & \textbf{29.40} & \textbf{34.80} & \textbf{34.20} & \textbf{35.80} & \textbf{35.40} & \textbf{58.25} & \textbf{59.92} & \textbf{59.92} & \textbf{59.38} & \textbf{60.00} \\
    \cmidrule(lr){2-13}
    &  \multirow{2}{*}{\shortstack{Q-value\\rankings}} & L2R & 27.20  & 27.80  & 27.80  & 29.60  & 30.20  & 54.83  & 56.83  & 55.67  & 54.42  & 54.75 \\
    &  & Ours & \textbf{27.80} & \textbf{30.60} & \textbf{32.20} & \textbf{31.20} & \textbf{30.40} & \textbf{56.71} & \textbf{58.50} & \textbf{58.21} & \textbf{58.04} & \textbf{57.50} \\
    \midrule\midrule
    \multirow{6}{*}{\shortstack{Llama-3-\\70B-Instruct}} 
    &  \multirow{2}{*}{BCE} & L2R & 39.80  & 39.60  & 39.20  & 41.40  & 41.20  & 70.96  & 70.25  & 69.38  & 68.08  & 65.75 \\
    &  & Ours & \textbf{44.60} & \textbf{43.20} & \textbf{43.00} & \textbf{43.60} & \textbf{43.00} & \textbf{72.04} & \textbf{71.25} & \textbf{71.25} & \textbf{70.79} & \textbf{70.04} \\
    \cmidrule(lr){2-13}
    &  \multirow{2}{*}{MSE} & L2R & 42.00  & 43.60  & 42.60  & 42.40  & 42.00  & 70.96  & 71.08  & 70.75  & 69.50  & 68.04 \\
    &  & Ours & \textbf{47.80} & \textbf{48.00} & \textbf{46.60} & \textbf{49.20} & \textbf{49.00} & \textbf{71.58} & \textbf{71.54} & \textbf{71.50} & \textbf{71.21} & \textbf{70.21} \\
    \cmidrule(lr){2-13}
    &  \multirow{2}{*}{\shortstack{Q-value\\rankings}} & L2R & 40.80  & 41.40  & 38.00  & 40.20  & 42.40  & 71.83  & 70.92  & 70.67  & 70.17  & 70.17 \\
    &  & Ours & \textbf{44.80} & \textbf{44.00} & \textbf{43.00} & \textbf{42.40} & \textbf{45.40} & \textbf{72.38} & \textbf{71.92} & \textbf{72.04} & \textbf{71.12} & \textbf{70.29} \\
    \bottomrule
    \end{tabular}
    \caption{\textbf{Deepseek-Math-7B} results measured by Best-of-N (BON@$n$) accuracy across two benchmarks, three PRM objectives and three sampling policies.}
    \label{tab:deepseek results}
\end{table*}

\section{Sensitivity Analysis of Aggregation Operators}
\label{sec:aggregation_analysis}
We investigate the sensitivity of BiPRM to four trajectory-level aggregation operators: product ($\prod$), minimum ($\min$), maximum ($\max$), and arithmetic mean ($\text{mean}$). 
Table~\ref{tab:aggregation} presents the detailed breakdown of BON@$N$ scores across all experimental configurations.
Quantitative analysis reveals that the aggregation strategy significantly influences performance.
The poor performance of $\max$ aligns with intuition, as a single high-quality step cannot compensate for logical errors elsewhere in the trajectory.
Conversely, the success of both $\text{mean}$ and $\min$ suggests that considering the overall quality or the worst-case step are both effective strategies.
Despite the slight empirical edge of $\text{mean}$, we adopt $\min$ as the default strategy. This decision aligns with the "weakest link" principle of mathematical reasoning where a single error invalidates the entire proof, thereby ensuring theoretical consistency with the logical nature of the task and prior literature~\cite{wang2024math, li2024process}.

\begin{table*}[!ht]
    \setlength{\tabcolsep}{1.5mm}
    \begin{tabularx}{\textwidth}{X|cccc}
    \toprule
    \textbf{Question:} Determine the maximum value of $c$ where $-2$ is within the range of the function $f(x) = x^2 + 3x + c$. & \multirow{2}{*}{Label} & \multirow{2}{*}{\shortstack{Bi-\\PRM}} & \multirow{2}{*}{\shortstack{L2R\\-PRM}} & \multirow{2}{*}{\shortstack{R2L\\-PRM}} \\
    \midrule
    \textbf{Step 1:} The range of a quadratic function $f(x) = ax^2 + bx + c$ is determined by the vertex of the parabola. & True & 1.00 & 1.00 & 1.00 \\
    \textbf{Step 2:} The $x$-coordinate of the vertex is $-\frac{b}{2a}$, and the $y$-coordinate is $c - \frac{b^2}{4a}$. & True & 1.00 & 0.91 & 1.00 \\
    \textbf{\textcolor{deepred}{Step 3:}} For the given function $f(x) = x^2 + 3x + c$, the vertex is $(-\frac{3}{2}, c - \frac{3^2}{4}) = (-1.5, c - \textbf{\textcolor{deepred}{4.5}})$. & False & 0.05 & 0.46 & 0.05 \\
    \textbf{Step 4:} Since $-2$ is within the range, we have $-2 \le c - 4.5$, which implies $c \ge 2.5$. & False & 0.04 & 0.01 & 0.09 \\
    \textbf{Step 5:} Therefore, the maximum value of $c$ is $\boxed{2.5}$. & False & 0.00 & 0.00 & 0.03 \\
    \bottomrule
    \end{tabularx}
    \caption{A success case study involving a calculation error. The critical error occurs in \textbf{Step 3} (highlighted in red), where the $y$-coordinate of the vertex is miscalculated. BiPRM effectively utilizes the R2L signal to correct the ambiguous evaluation of the L2R baseline.}
    \label{tab:case-good}
\end{table*}

\begin{table*}[!ht]
    \setlength{\tabcolsep}{1.5mm}
    \begin{tabularx}{\textwidth}{X|cccc}
    \toprule
    \textbf{Question:} If the area of a circle is $M\text{cm}^2$  and the circumference is $N$ cm , with $M/N = 20$, what is the radius of the circle in cm? & \multirow{2}{*}{Label} & \multirow{2}{*}{\shortstack{Bi-\\PRM}} & \multirow{2}{*}{\shortstack{L2R\\-PRM}} & \multirow{2}{*}{\shortstack{R2L\\-PRM}} \\
    \midrule
    \textbf{Step 1:} The area of a circle is given by the formula $A = \pi r^2$, and the circumference is given by the formula $C = 2\pi r$. & \raisebox{-13pt}{True} & \raisebox{-13pt}{1.00} & \raisebox{-13pt}{1.00} & \raisebox{-13pt}{1.00} \\
    \textbf{Step 2:} We are given that $M = A$ and $N = C$, and $M/N = 20$. & True & 1.00 & 1.00 & 1.00 \\
    \textbf{Step 3:} Substituting the formulas, we have $\pi r^2 / (2\pi r) = 20$. & False & 1.00 & 1.00 & 1.00 \\
    \textbf{\textcolor{deepred}{Step 4:}} Simplifying, we get \textcolor{deepred}{$\bm {r^2 / r = 20}$}.  & False & 0.18 & 0.01 & 0.63 \\
    \textbf{Step 5:} Dividing both sides by $r$, we get $r = 20$. The answer is: \boxed{20}. & False & 0.00 & 0.00 & 0.01 \\
    \bottomrule
    \end{tabularx}
    \caption{A failure case study demonstrating the limitation of R2L-PRM. The critical error arises during the algebraic simplification transition between Step 3 and Step 4. Since the derivation from Step 4 onwards maintains internal consistency, the retrospective view of the R2L stream fails to identify the preceding disconnect.}
    \label{tab:case-bad}
\end{table*}

\begin{table*}[!ht]
    \centering
    \small
    \begin{tabular}{ccc|ccc|ccc|c}
    \toprule
    \multirow{3}{*}{\shortstack{Sampling\\Policy}} & \multirow{3}{*}{\shortstack{Backbone}} & \multirow{3}{*}{\shortstack{Aggregation\\Operator}} & \multicolumn{3}{c|}{Dataset: MATH500} & \multicolumn{3}{c|}{Dataset: GSM-Plus} & \multirow{3}{*}{\shortstack{Avg}} \\
    &  &  & \multirow{2}{*}{BCE} & \multirow{2}{*}{MSE} & \multirow{2}{*}{\shortstack{Q-value\\rankings}} & \multirow{2}{*}{BCE} & \multirow{2}{*}{MSE} & \multirow{2}{*}{\shortstack{Q-value\\rankings}} &  \\
    & & & & & & & & &  \\
    \toprule
    \multirow{12}{*}{\shortstack{MetaMath-\\Mistral-7B}}
    &  \multirow{4}{*}{Rho-Math-1B} & $\prod$ & 24.52 & 24.28 & 25.08 & 45.87 & 45.71 & 47.50 & 35.49 \\
    &  & min & 23.56 & 24.96 & 26.04 & 45.89 & 45.28 & 47.96 & 35.62 \\
    &  & max & 22.96 & 24.64 & 26.60 & 46.76 & 49.61 & 49.32 & \textbf{36.65} \\
    &  & mean & 24.64 & 25.00 & 26.68 & 45.44 & 46.75 & 48.61 & \underline{36.19} \\
    \cmidrule(lr){2-10}
    &  \multirow{4}{*}{Qwen2.5-Math-1.5B} & $\prod$ & 38.76 & 39.64 & 40.00 & 54.88 & 57.20 & 58.14 & 48.10 \\
    &  & min & 38.56 & 39.80 & 40.24 & 55.02 & 58.11 & 58.87 & \underline{48.43} \\
    &  & max & 30.64 & 36.24 & 41.44 & 51.02 & 55.26 & 58.73 & 45.55 \\
    &  & mean & 38.64 & 39.52 & 40.56 & 54.99 & 58.29 & 58.69 & \textbf{48.45} \\
    \cmidrule(lr){2-10}
    &  \multirow{4}{*}{Deepseek-Math-7B} & $\prod$ & 36.20 & 37.36 & 32.72 & 56.59 & 58.99 & 56.69 & 46.43 \\
    &  & min & 36.44 & 37.40 & 34.48 & 56.96 & 59.52 & 57.13 & \underline{46.99} \\
    &  & max & 32.80 & 34.44 & 33.36 & 53.13 & 57.84 & 56.97 & 44.76 \\
    &  & mean & 37.00 & 37.76 & 34.28 & 57.13 & 59.76 & 57.43 & \textbf{47.23} \\
    \midrule
    \multirow{12}{*}{\shortstack{Muggle-\\Math-13B}} 
    &  \multirow{4}{*}{Rho-Math-1B} & $\prod$ & 21.36 & 20.28 & 21.44 & 41.30 & 44.46 & 45.55 & \underline{32.40} \\
    &  & min & 21.00 & 19.52 & 20.56 & 42.89 & 44.02 & 45.09 & 32.18 \\
    &  & max & 18.52 & 18.68 & 19.64 & 41.38 & 45.92 & 46.28 & 31.74 \\
    &  & mean & 21.28 & 20.40 & 20.92 & 41.33 & 45.12 & 45.53 & \textbf{32.43} \\
    \cmidrule(lr){2-10}
    &  \multirow{4}{*}{Qwen2.5-Math-1.5B} & $\prod$ & 33.96 & 34.88 & 35.60 & 56.17 & 57.13 & 57.04 & 45.80 \\
    &  & min & 33.92 & 34.96 & 36.08 & 56.33 & 57.43 & 58.46 & \textbf{46.20} \\
    &  & max & 25.00 & 32.92 & 37.16 & 49.81 & 54.17 & 57.08 & 42.69 \\
    &  & mean & 33.68 & 35.24 & 36.80 & 55.92 & 57.31 & 57.67 & \underline{46.10} \\
    \cmidrule(lr){2-10}
    &  \multirow{4}{*}{Deepseek-Math-7B} & $\prod$ & 32.84 & 33.64 & 30.08 & 59.66 & 58.53 & 56.62 & 45.23 \\
    &  & min & 32.64 & 33.92 & 30.44 & 59.57 & 59.49 & 57.79 & \textbf{45.64} \\
    &  & max & 25.16 & 30.92 & 31.16 & 52.97 & 56.78 & 56.71 & 42.28 \\
    &  & mean & 32.60 & 33.36 & 30.68 & 59.12 & 59.63 & 57.25 & \underline{45.44} \\
    \midrule
    \multirow{12}{*}{\shortstack{Llama-3-\\70B-Instruct}} 
    &  \multirow{4}{*}{Rho-Math-1B} & $\prod$ & 35.84 & 38.76 & 38.80 & 66.85 & 68.03 & 67.86 & 52.69 \\
    &  & min & 37.32 & 38.24 & 39.08 & 67.73 & 68.64 & 68.92 & \underline{53.32} \\
    &  & max & 36.44 & 37.00 & 37.72 & 67.53 & 69.27 & 69.73 & 52.95 \\
    &  & mean & 35.48 & 38.96 & 38.92 & 68.19 & 69.48 & 69.40 & \textbf{53.41} \\
    \cmidrule(lr){2-10}
    &  \multirow{4}{*}{Qwen2.5-Math-1.5B} & $\prod$ & 48.44 & 46.52 & 48.60 & 70.47 & 70.61 & 70.66 & 59.22 \\
    &  & min & 48.48 & 46.68 & 48.92 & 70.44 & 70.93 & 71.20 & \underline{59.44} \\
    &  & max & 45.40 & 46.16 & 48.48 & 70.55 & 70.62 & 71.39 & 58.77 \\
    &  & mean & 48.28 & 47.72 & 48.92 & 70.73 & 71.20 & 71.15 & \textbf{59.67} \\
    \cmidrule(lr){2-10}
    &  \multirow{4}{*}{Deepseek-Math-7B} & $\prod$ & 42.92 & 46.96 & 43.76 & 70.48 & 71.18 & 70.89 & 57.70 \\
    &  & min & 43.48 & 48.12 & 43.92 & 71.07 & 71.21 & 71.55 & \textbf{58.23} \\
    &  & max & 40.72 & 47.28 & 44.00 & 70.32 & 71.26 & 71.62 & 57.53 \\
    &  & mean & 41.88 & 48.48 & 44.44 & 70.88 & 71.27 & 71.95 & \underline{58.15} \\
    \toprule
    \end{tabular}
    \caption{Sensitivity analysis of different trajectory-level aggregation operators across diverse backbones and objectives. The metrics reported are the average BON@$N$ scores. Across all configurations, the overall average scores are: mean (47.45) $>$ min (47.34) $>$ $\prod$ (47.01) $>$ max (45.88).
    }
    \label{tab:aggregation}
\end{table*}

\end{document}